\documentclass[runningheads]{llncs}

\usepackage{eccv}

\usepackage{hyperref}

\usepackage{bbm}
\usepackage{pgffor}
\usepackage{xcolor}
\usepackage{xparse}
\usepackage{amsmath}
\usepackage{siunitx}
\usepackage{amssymb}
\usepackage{graphicx}
\usepackage{etoolbox}
\usepackage{orcidlink}
\usepackage[T1]{fontenc}
\usepackage[utf8]{inputenc}
\usepackage[english]{babel}
\usepackage[accsupp]{axessibility}
\usepackage[autostyle=true]{csquotes}
\usepackage[nolist, nohyperlinks]{acronym}
\usepackage[capitalize, sort, english]{cleveref}

\usepackage{tikz}
\usetikzlibrary{
  arrows.meta,
  positioning,
  fit,
  backgrounds,
  calc,
  decorations.pathreplacing
}
 
\usepackage{booktabs}
\usepackage{tabularx}
\usepackage{multirow}
\usepackage{array}

\newcolumntype{R}{>{\hsize=.5\hsize\raggedleft\arraybackslash}X}

\crefname{equation}{Eq.}{Eqs.}
\Crefname{equation}{Equation}{Equations}
\crefformat{equation}{Eq.~#2#1#3}
\Crefformat{equation}{Equation~#2#1#3}

\newcommand{\etal}{\textit{et~al.}}

\begin{acronym}
\acro{AAT}{\textit{Art \& Architecture Thesaurus}}
\acro{AI}{\textit{Artificial Intelligence}}
\acro{CRM}{\textit{Conceptual Reference Model}}
\acro{CV}{\textit{Computer Vision}}
\acro{LOD}{\textit{Linked Open Data}}
\acro{MAE}{\textit{Mean Absolute Error}}
\acro{PCA}{\textit{Principal Component Analysis}}
\acro{RMSE}{\textit{Root Mean Squared Error}}
\acro{ULAN}{\textit{Union List of Artist Names}}
\acro{VLM}{\textit{Vision--Language Model}}
\end{acronym}

\begin{document}

\title{Uncertainty-Aware Art-Historical Dating \\with Vision--Language Models}

\titlerunning{Uncertainty-Aware Art-Historical Dating}

\author{
Stefanie Schneider\orcidlink{0000-0003-4915-6949} \and
Peter Bell\orcidlink{0000-0003-4415-7408}
}

\authorrunning{S.~Schneider and P.~Bell}

\institute{Marburg University, Biegenstraße 11, 35037 Marburg, Germany
\email{\{stefanie.schneider,peter.bell\}@uni-marburg.de}}

\maketitle

\begin{abstract}
    Museum and archival datasets do not mirror historical artistic production, but materialize the contingent histories of collecting, preservation, cataloging, and digitization.
    This has direct consequences for interpreting pretrained image representations: they may appear to encode historical time while actually encoding the institutional conditions under which objects become visible as data.
    We describe this phenomenon as \textit{temporal entanglement} and investigate it by formulating artwork dating as an uncertainty-aware regression task over frozen image embeddings.
    We evaluate several pretrained vision models on a temporally controlled Wikidata corpus of artworks.
    Our results show that these models contain usable temporal information, with \acp{VLM} outperforming purely visual self-supervised baselines.
    However, a qualitative analysis indicates that this temporal knowledge is shaped by various biases.
    
\end{abstract}

\acresetall


\section{Introduction}
\label{sec:introduction}

Museum and archival datasets do not mirror historical artistic production~\cite{kizhner2021}.
Rather, they are produced through historically contingent processes of acquisition, selection, preservation, and digitization~\cite{drucker2011}.
This has significant implications for the use of pretrained image-representation models in cultural-heritage applications, including \acp{VLM} such as SigLIP2~\cite{siglip2}, which are now being used increasingly for artwork retrieval~\cite{offert2024,springstein2021} and classification~\cite{baldrati2022,conde2021}.
At first glance, such models appear to recover historically meaningful visual regularities; their representations seem to \enquote*{know} something about when an artwork was made.
Indeed, prior work has shown that vision models can, to some extent, predict style, period, or related attributes~\cite{kim2025,strafforello2025}.
The question, however, is what kind of time is being encoded.
We argue that the temporal signal in these models should not be interpreted as a direct trace of artistic production.
Rather, it should be understood as a mediated signal in which the technical and institutional processes by which objects become data may become intertwined.
We call this mismatch \textit{temporal entanglement}.
This term describes the divergence between a model's apparent historical temporality and the institutional conditions under which that temporality is produced.
Temporal entanglement is not separate from data bias, but a specifically temporal manifestation of it: institutional selection and mediation help organize the chronological structure that models appear to learn.
We use the term \textit{entanglement} rather than \textit{confounding}, since our experiments are diagnostic and not designed to identify a formal causal relationship.

In this paper, we employ temporal estimation both as a predictive task and a diagnostic tool for investigating this entanglement.
Rather than asking only whether pretrained representations contain usable temporal information, we ask what kind of temporality they make learnable.
Since temporal estimation presupposes a spatialization of historical time, dates become inferable when historical distance is translated into distance between image representations.
Following Offert~\cite{offert2023}, we therefore approach pretrained vision models not as neutral feature extractors, but as technical systems that operationalize a specific \enquote{concept of history}---a patterned way in which the past becomes available, ordered, and made actionable.
This perspective requires a methodology that respects the uncertainty of art-historical dating, which is often approximate, interval-based, and contested.
Treating creation dates as exact point labels, as in prior work~\cite{fernando2014,li2018,strafforello2025}, therefore risks eliminating historical ambiguity.
Our contributions can be summarized as follows:
(1)~We introduce \textit{temporal entanglement} as a diagnostic framework for analyzing pretrained image representations in cultural heritage, arguing that temporal knowledge encoded in such representations should be interpreted in relation to the institutional histories through which objects become visible as data, rather than as direct evidence that models recover creation time itself.
(2)~Our approach is formulated as an uncertainty-aware temporal-estimation task over frozen pretrained image embeddings, combining quantile regression with split-conformal calibration to produce calibrated temporal intervals.
(3)~In this context, we construct a temporally controlled Wikidata corpus that preserves interval-based creation-date metadata and balances the sample across twelve fifty-year strata and four major object types.
(4)~We evaluate several \acp{VLM} and vision-only models quantitatively and qualitatively, showing that chronological structure coexists with systematic biases.

The remainder of this paper is organized as follows:
First, in \cref{sec:related-work}, we situate our approach in the context of existing work.
\cref{sec:temporal-estimation} then presents the proposed methodology for temporal estimation, and \cref{sec:dataset} describes the construction of a balanced, temporally controlled Wikidata corpus.
\cref{sec:experiments} details the experimental setup, reports the quantitative results, and discusses common failure modes through a qualitative error analysis.
Finally, \cref{sec:conclusions} summarizes the paper's contributions and outlines directions for future work.

\section{Related Work}
\label{sec:related-work}


\subsection{Biases in Cultural-Heritage Datasets}
\label{sec:rw-data-biases}

Bias in computational systems is not only a property of model output, but a socio-technical phenomenon that can emerge before, during, and after system design.
Friedman and Nissenbaum's distinction between pre-existing, technical, and emergent bias~\cite{friedman1996} is especially useful for cultural-heritage applications because it separates different moments at which historical time becomes computationally mediated.
Temporal entanglement can be understood as a setting in which these forms of bias become organized chronologically:
\textit{pre-existing} bias enters through the histories of collecting that influence which objects become available as data;
\textit{technical bias} enters when uncertain dates, periods, or stylistic categories are converted into machine-learning targets;
\textit{emergent bias} appears when models trained on one institutional or geographic corpus are used to interpret objects from different historical contexts. 
Recent machine-learning literature similarly treats bias as distributed across the data and model life cycle~\cite{suresh2021}.
In art-historical settings, however, these problems are intensified because museum and archival datasets are not representative samples of artistic production, but selective records of institutional visibility~\cite{drucker2011,kizhner2021}.
Accordingly, a model that appears to encode historical time may also encode the conditions under which certain objects have been collected and digitized.

\subsection{Temporal Signals in Visual Representations}
\label{sec:rw-temporal-estimation}

Prior work has already shown that visual representations encode temporal information, although not necessarily the same kind of temporality.
In historical photography, time is usually operationalized as the acquisition date of the image.
Fernando \etal~\cite{fernando2014} classify historical photographs by decade using hand-crafted color descriptors. 
Müller \etal~\cite{muller2017} extend this line of work with convolutional regression and classification models for photographs from the twentieth century.
Molina \etal~\cite{molina2021} formulate dating as a retrieval problem, learning embeddings in which visual and temporal proximity are aligned.
More recently, Barancová \etal~\cite{barancova2023} show that OpenCLIP~\cite{openclip} performs weakly, and in temporally biased ways, in a zero-shot setting on historical press photographs. 
For artworks and cultural-heritage objects, however, temporality is usually defined not by the moment an image was captured, but by creation date, period, movement, or style---categories that are themselves approximate and historically mediated.
Li \etal~\cite{li2018}, for instance, combine color, shape, and convolutional features to support the dating of murals.
Strafforello \etal~\cite{strafforello2025} evaluate whether general-purpose \acp{VLM} can infer style, author, and period, finding that such models recover some art-historical attributes, but perform unevenly across prompts, artists, styles, and datasets.
Kim \etal~\cite{kim2025}, using Stable Diffusion~\cite{stable-diffusion}, argue that semantic information can distinguish periods, styles, and artists more effectively than formal properties alone.
These works show that pretrained visual models can encode art-historical signals, but they also suggest that such signals are task-dependent and vulnerable to dataset composition.
Most of them, however, represent time either as a classification label or as a point-valued regression target~\cite{fernando2014,li2018,strafforello2025}.
Approximate date ranges are thereby often collapsed into exact labels.

Our approach differs from this prior work in two respects.
First, rather than relying on established benchmarks such as WikiArt~\cite{kim2025,strafforello2025}, we construct a Wikidata-derived corpus that preserves interval-based and vague temporal metadata.
Second, rather than using uncertainty as an auxiliary confidence score, we make it part of the temporal prediction itself.
By combining quantile regression with conformal calibration, the model produces both a central date estimate and a calibrated temporal interval.
In so doing, temporal estimation allows us to examine when pretrained image representations encode historically meaningful temporal structure, and when their predictions appear to align with, amplify, or make visible the biases of the collections from which visual data are assembled.

\section{Uncertainty-aware Temporal Estimation}
\label{sec:temporal-estimation}

Building on the capacity of pretrained visual representations to encode visual-semantic regularities across large-scale image collections, we formulate temporal estimation as a supervised regression problem over frozen pretrained image embeddings.
The following subsections first define the task formally (\cref{sec:task-definition}) and then describe the methodological pipeline (\cref{sec:methodology}).

\subsection{Task Definition}
\label{sec:task-definition}

Each artwork $A$ consists of an image $I_A$, descriptive metadata $M_A$, and, where available, a creation date $D_A$. 
Since art-historical dates are often approximate or expressed as ranges, we represent $D_A$ as an interval $D_A = \left[y_A^{\mathrm{start}}, y_A^{\mathrm{end}}\right]$, where $y_A^{\mathrm{start}}$ and $y_A^{\mathrm{end}}$ denote the earliest and latest plausible creation years, respectively.
This representation follows established approaches to uncertain temporal metadata based on temporal bounds.\footnote{See, e.g., the CIDOC \ac{CRM} \cite{cidoc-crm} that supports statements about minimum and maximum temporal extents.} 
Given $I_A$, we predict (1)~a creation-year estimate $\hat{y}_A$ and (2)~an uncertainty-aware predictive interval $\hat{D}_A = \left[\hat{y}_A^{\mathrm{lower}}, \hat{y}_A^{\mathrm{upper}}\right]$.
The point estimate $\hat{y}_A$ represents the model's central temporal attribution, while $\hat{D}_A$ describes the range of plausible creation dates inferred from the visual evidence.
Formally, a pretrained visual encoder $\phi$ produces an image representation $x_A=\phi(I_A)$, from which $f_\theta$ predicts: 
\begin{equation}
f_\theta(x_A) = \left(\hat{y}_A, \hat{D}_A\right).
\end{equation}
For example, an artwork cataloged as \enquote{1620--1630} is represented by the interval $D_A = [1620, 1630]$.
The model is then expected to infer, from the artwork image alone, both a plausible creation-date estimate and a temporal interval that reflects the uncertainty of this inference.


\definecolor{imgblue}{HTML}{DDEBF7}
\definecolor{modelgreen}{HTML}{E2F0D9}
\definecolor{outgray}{HTML}{EDEDED}
\definecolor{linegray}{HTML}{404040}


\newcommand{\LandscapePanel}{%
  \draw[fill=white, draw=linegray!35, line width=0.4pt]
    (0.18,0.18) rectangle (1.82,1.27);

  \begin{scope}
    \clip (0.18,0.18) rectangle (1.82,1.27);

    \fill[linegray!6] (0.18,0.18) rectangle (1.82,1.27);

    \fill[linegray!35] (1.45,1.02) circle (0.11);

    \fill[linegray!18]
      (0.2,0.1) -- (0.62,0.88) -- (0.96,0.44)
      -- (1.18,0.72) -- (1.82,0.25) -- cycle;

    \fill[linegray!30]
      (0,0.1) .. controls (0.55,0.42) and (1.10,0.36) ..
      (1.82,0.58) -- (1.82,0.18) -- cycle;
  \end{scope}

  \draw[linegray!70, line width=0.45pt]
    (0.18,0.18) rectangle (1.82,1.27);
}

\newcommand{\ArtworkRecordIcon}{%
  \begin{tikzpicture}[scale=0.50]
    \draw[fill=white, draw=linegray, thick] (0,0) rectangle (2.0,2.8);
    \draw[linegray] (0.25,2.45) -- (1.75,2.45);
    \draw[linegray] (0.25,2.15) -- (1.75,2.15);
    \draw[linegray] (0.25,1.85) -- (1.35,1.85);
    \draw[linegray] (0.25,1.55) -- (1.65,1.55);

    \begin{scope}[
      shift={(1.0,0.725)},
      scale=0.88,
      shift={(-1.0,-0.725)}
    ]
      \LandscapePanel
    \end{scope}
  \end{tikzpicture}%
}

\newcommand{\ImageIcon}{%
  \begin{tikzpicture}[x=1cm,y=1cm,scale=0.48,baseline=-0.5ex]
    \LandscapePanel
  \end{tikzpicture}%
}

\newcommand{\EmbeddingIcon}{%
  \begin{tikzpicture}[x=1cm,y=1cm,baseline=-0.5ex]
    \path[use as bounding box] (0.25,0.15) rectangle (2.05,0.62);

    \draw[linegray!25, line width=0.5pt]
      (0.28,0.30)
      .. controls (0.65,0.58) and (1.05,0.18) .. (1.42,0.42)
      .. controls (1.65,0.57) and (1.88,0.47) .. (2.02,0.35);

    \foreach \x/\y in {
      0.42/0.45,
      0.56/0.35,
      0.68/0.56,
      0.82/0.22,
      0.96/0.43,
      1.28/0.50,
      1.44/0.33,
      1.60/0.58,
      1.74/0.40,
      1.90/0.25,
      2.00/0.48
    }{
      \fill[linegray!45] (\x,\y) circle (0.026);
    }

    \draw[linegray, line width=0.55pt, fill=white] (1.18,0.33) circle (0.075);
    \fill[linegray] (1.18,0.33) circle (0.032);
  \end{tikzpicture}%
}

\newcommand{\CatalogIntervalIcon}{%
  \begin{tikzpicture}[x=1cm,y=1cm,baseline=-0.5ex]
    \draw[linegray!35, line width=0.6pt] (0,0) -- (2.0,0);
    \draw[linegray, line width=1.2pt] (0.35,0) -- (1.65,0);

    \draw[linegray, thick] (0.35,-0.13) -- (0.35,0.13);
    \draw[linegray, thick] (1.65,-0.13) -- (1.65,0.13);
  \end{tikzpicture}%
}

\newcommand{\MidpointTargetIcon}{%
  \begin{tikzpicture}[x=1cm,y=1cm,baseline=-0.5ex]
    \draw[linegray!35, line width=1.2pt] (0.35,0) -- (1.65,0);

    \draw[linegray!60, thick] (0.35,-0.13) -- (0.35,0.13);
    \draw[linegray!60, thick] (1.65,-0.13) -- (1.65,0.13);

    \fill[linegray] (1.0,0) circle (0.075);
  \end{tikzpicture}%
}

\newcommand{\PredictedYearIcon}{%
  \begin{tikzpicture}[x=1cm,y=1cm,baseline=-0.5ex]
    \draw[white, thick] (0.35,-0.13) -- (0.35,0.13);
    \draw[white, thick] (1.65,-0.13) -- (1.65,0.13);

    \draw[linegray!35, line width=0.6pt] (0,0) -- (2.0,0);

    \draw[linegray, line width=0.8pt, fill=white] (1.0,0) circle (0.075);
  \end{tikzpicture}%
}

\newcommand{\PredictiveIntervalIcon}{%
  \begin{tikzpicture}[x=1cm,y=1cm,baseline=-0.5ex]
    \draw[linegray!35, line width=0.6pt] (0,0) -- (2.0,0);
    \draw[linegray, line width=1.2pt] (0.35,0) -- (1.65,0);

    \draw[linegray, thick] (0.35,-0.13) -- (0.35,0.13);
    \draw[linegray, thick] (0.35,-0.13) -- (0.45,-0.13);
    \draw[linegray, thick] (0.35,0.13) -- (0.45,0.13);

    \draw[linegray, thick] (1.65,-0.13) -- (1.65,0.13);
    \draw[linegray, thick] (1.55,-0.13) -- (1.65,-0.13);
    \draw[linegray, thick] (1.55,0.13) -- (1.65,0.13);
  \end{tikzpicture}%
}


\tikzset{
  method pipeline/.style={
    font=\scriptsize,
    node distance=0.5cm and 0.75cm
  },
  pipeline arrow/.style={
    -{Latex[length=2mm]},
    thick,
    draw=linegray
  },
  pipeline soft arrow/.style={
    -{Latex[length=2mm]},
    thick,
    dashed,
    draw=linegray
  },
  pipeline box/.style={
    draw=linegray,
    thick,
    rounded corners=2pt,
    align=center,
    inner sep=4pt,
    minimum height=0.5cm,
    text width=2.6cm
  },
  pipeline group box/.style={
    draw=linegray!50,
    thick,
    dotted,
    rounded corners=2pt,
    inner sep=6pt
  },
  pipeline group label/.style={
    fill=white,
    inner sep=2pt
  }
}

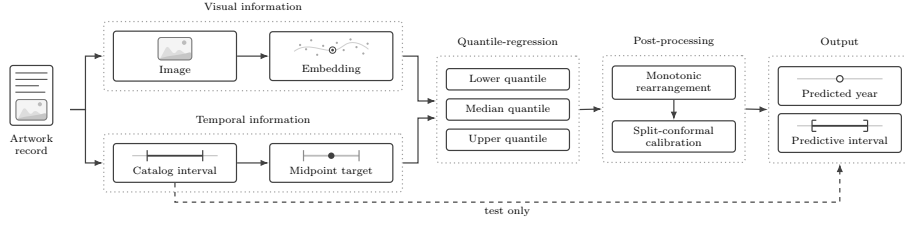
\begin{figure}[t]
\centering
\scriptsize
\resizebox{\linewidth}{!}{%
\begin{tikzpicture}[method pipeline]


\node[
  pipeline box,
  draw=none,
  minimum height=2.65cm,
  text width=1.5cm
] (record) at (0,0) {
  \ArtworkRecordIcon\\[0.1cm]
  Artwork\\record
};


\node[pipeline box, right=1cm of record, yshift=1.25cm] (image) {
  \ImageIcon\\[-0.4em]
  Image
};

\node[pipeline box, right=of image] (embedding) {
  \EmbeddingIcon\\[-0.5em]
  Embedding
};

\node[pipeline group box, fit=(image)(embedding)] (visual-group) {};
\node[pipeline group label, anchor=north, yshift=0.5cm]
  at (visual-group.north) {Visual information};


\node[pipeline box, right=1cm of record, yshift=-1.25cm] (interval) {
  \CatalogIntervalIcon\\[0.1cm]
  Catalog interval
};

\node[pipeline box, right=of interval] (target) {
  \MidpointTargetIcon\\[0.1cm]
  Midpoint target
};

\node[pipeline group box, fit=(interval)(target)] (temporal-group) {};
\node[pipeline group label, anchor=north, yshift=0.5cm]
  at (temporal-group.north) {Temporal information};


\node[pipeline box, anchor=west] (quantile-median)
  at ($(visual-group.east |- record.center) + (1cm,0)$)
  {Median quantile};

\node[pipeline box, above=0.18cm of quantile-median] (quantile-lower)
  {Lower quantile};

\node[pipeline box, below=0.18cm of quantile-median] (quantile-upper)
  {Upper quantile};

\node[
  pipeline group box,
  fit=(quantile-lower)(quantile-median)(quantile-upper),
  minimum height=2.5cm
] (quantile-group) {};

\node[pipeline group label, anchor=north, yshift=0.5cm]
  at (quantile-group.north) {Quantile-regression};

\coordinate (qvis-in)
  at ($(quantile-group.west |- quantile-median.west) + (0,0.2cm)$);

\coordinate (qtemp-in)
  at ($(quantile-group.west |- quantile-median.west) + (0,-0.2cm)$);


\coordinate (post-center)
  at ($(quantile-group.east |- record.center) + (0.75cm,0)$);

\node[pipeline box, anchor=west] (sort)
  at ($(post-center) + (0,0.63cm)$)
  {Monotonic\\rearrangement};

\node[pipeline box, anchor=west] (conf)
  at ($(post-center) + (0,-0.63cm)$)
  {Split-conformal\\calibration};

\node[
  pipeline group box,
  fit=(sort)(conf),
  minimum height=2.5cm
] (post-group) {};

\node[pipeline group label, anchor=north, yshift=0.5cm]
  at (post-group.north) {Post-processing};


\coordinate (output-center)
  at ($(post-group.east |- record.center) + (0.75cm,0)$);

\node[pipeline box, anchor=west] (predicted-year)
  at ($(output-center) + (0,0.55cm)$)
  {
    \PredictedYearIcon\\[0.1cm]
    Predicted year
  };

\node[pipeline box, anchor=west] (predicted-interval)
  at ($(output-center) + (0,-0.55cm)$)
  {
    \PredictiveIntervalIcon\\[0.1cm]
    \phantom{y}Predictive interval\phantom{y}
  };

\node[
  pipeline group box,
  fit=(predicted-year)(predicted-interval),
  minimum height=2.5cm
] (output-group) {};

\node[pipeline group label, anchor=north, yshift=0.5cm]
  at (output-group.north) {Output};


\draw[pipeline arrow] (record.east) -- ++(0.36,0) |- (visual-group.west);
\draw[pipeline arrow] (record.east) -- ++(0.36,0) |- (temporal-group.west);

\draw[pipeline arrow] (image) -- (embedding);
\draw[pipeline arrow] (interval) -- (target);

\draw[pipeline arrow]
  (visual-group.east) -- ++(0.35,0) |- (qvis-in);

\draw[pipeline arrow]
  (temporal-group.east) -- ++(0.35,0) |- (qtemp-in);

\draw[pipeline arrow] (sort) -- (conf);

\draw[pipeline arrow] (quantile-group.east) -- (post-group.west);
\draw[pipeline arrow] (post-group.east) -- (output-group.west);

\coordinate (test-route-left) at ($(interval.south) + (0,-0.45cm)$);
\coordinate (test-route-right) at (output-group.south |- test-route-left);

\draw[pipeline soft arrow]
  (interval.south) -- (test-route-left)
  -- node[midway, below] {test only}
  (test-route-right)
  -- (output-group.south);

\end{tikzpicture}
}%

\caption{
    Pipeline for uncertainty-aware temporal estimation of artwork creation dates.
    From an artwork record, we extract visual features from each image and convert the temporal metadata into scalar midpoint targets for training.
    Quantile-regression models then estimate the conditional quantiles of the creation year, outputting a central prediction with a calibrated predictive interval.
}
\label{fig:methodological-pipeline}
\end{figure}

\subsection{Methodology}
\label{sec:methodology}

To address the task defined above, we employ a supervised estimation pipeline that combines visual representations with uncertainty-aware regression (\cref{fig:methodological-pipeline}).
The pipeline consists of four stages:
(1)~extracting image embeddings from artworks,
(2)~deriving scalar training targets from creation-date metadata,
(3)~fitting quantile-regression models, and
(4)~post-processing the resulting intervals.

\paragraph{Image Representation}

Given an artwork image $I_A$, we first extract a learned visual representation using a pretrained image encoder $\phi$.
The resulting embedding, $x_A \in \mathbb{R}^d$, serves as the input to all downstream estimators.
Since these embeddings are typically high-dimensional, we apply \ac{PCA} prior to regression.
This mitigates the risk of overfitting and standardizes the input dimensionality across encoder backbones.

\paragraph{Training Targets}

The creation-date label is derived from the temporal metadata associated with each artwork.
If the metadata specify a single creation year, that year is used directly as the regression target.
If the date is given as an interval $D_A = \left[y_A^{\mathrm{start}}, y_A^{\mathrm{end}}\right]$, we use its midpoint, $y_A$, as the scalar target for model fitting; the original interval boundaries are retained for evaluation.
The midpoint should not be interpreted as the true historical creation date, but rather as a pragmatic scalar approximation that enables supervised regression.
Importantly, the catalog interval is neither provided to the regressor as an input nor incorporated into the training loss: all quantile regressors are trained exclusively on the scalar midpoint targets. 
Accordingly, the median regressor is a point-only estimate, while the lower and upper regressors estimate conditional quantiles of the same scalar target distribution. 
The original catalog interval is used only at evaluation time, allowing us to distinguish prediction error relative to the operational midpoint from agreement with the temporal uncertainty encoded in the source metadata. 

\paragraph{Quantile Regression}

The estimator is formulated as a quantile-regression model.
Rather than learning only a single mapping from visual representation to creation year, the model estimates a set of conditional temporal quantiles:
\begin{equation}
    \hat{\mathcal{Q}}_A = {\hat{q}_{A,\tau} \mid \tau \in \mathcal{Q}},
\end{equation}
where $\mathcal{Q}$ denotes the set of quantile levels.
For each quantile $\tau$, a separate regression function is learned:
\begin{equation}
    \hat{q}_{A,\tau} = f_{\theta,\tau}(x_A).
\end{equation}
The median quantile $\hat{q}_{A,0.50}$ provides the central creation-date estimate, while the lower and upper quantiles define an uncertainty-aware temporal interval.
For a pair of quantile levels $(\tau_l,\tau_u)$ with $\tau_l < 0.50 < \tau_u$, the predicted interval is given by $\hat{D}_A = \left[\hat{q}_{A,\tau_l}, \hat{q}_{A,\tau_u}\right]$.
For each quantile level, the model parameters are optimized using the pinball loss~\cite{steinwart2011}:
\begin{equation}\label{eq:pinball-loss}
    \mathcal{L}_{\tau}(y_A,\hat{q}_{A,\tau}) =
    \max\left(\tau\left(y_A-\hat{q}_{A,\tau}\right),\left(\tau-1)(y_A-\hat{q}_{A,\tau}\right)\right).\footnote{We also tested an interval-aware pinball loss, but its zero-gradient region treated all in-interval predictions equally and weakened optimization.}
\end{equation}
This loss penalizes over- and under-estimation asymmetrically according to the selected quantile level.
The collection of quantile-specific regressors therefore approximates the conditional distribution of plausible creation dates given the visual representation of the artwork.
In other words, uncertainty is not added to the model after prediction, but is itself part of what the model learns to estimate.

\paragraph{Post-processing}

Because the quantile-specific regressors are fitted independently, their outputs may occasionally violate the expected monotonic ordering of temporal estimates \cite{chernozhukov2010}.
For each artwork, the predicted quantile values are therefore sorted in ascending order before intervals are constructed.
Intervals are then calibrated using split-conformal prediction~\cite{romano2019}.
For this purpose, a subset of the training data is held out as a calibration set, while the quantile-regression models are fitted on the remaining observations.
On the calibration examples, we compute conformity scores that measure the extent to which the target year falls outside the predicted interval.
The empirical quantile of these conformity scores is then used to enlarge the lower and upper bounds.
This step provides a conservative adjustment of the model's temporal intervals so that they better reflect the empirical distribution of prediction errors observed on held-out data.

\section{Dataset}
\label{sec:dataset}

The methodological framework described above requires a corpus in which images can be related to temporally explicit metadata; more precisely, it requires records whose dates can be represented, where necessary, as intervals rather than as single years.
We thus select Wikidata as the basis for our corpus.
Unlike a single museum collection or a benchmark-specific dataset, it offers a cross-institutional knowledge graph in which institutional, historical, and infrastructural asymmetries can be made visible and, consequently, analyzed.
At the same time, Wikidata is an aggregation layer rather than a uniformly curated museum catalog. 
Its statements may inherit heterogeneous institutional cataloging practices, and our extraction does not reconstruct the upstream provenance or curation history of every temporal statement. 
We therefore treat the resulting dates as operational metadata rather than ground-truth production dates; this heterogeneity is part of the motivation for evaluating agreement with catalog intervals and for interpreting model errors diagnostically. 

Using the Wikidata SPARQL endpoint,\footnote{\url{https://query.wikidata.org} (last accessed on \today).} we extract object records that are instances of either \enquote{visual artwork} (Wikidata item \texttt{Q4502142}) or \enquote{artwork series} (\texttt{Q15709879}).\footnote{See Schneider \etal\ \cite{schneider2026} for the respective SPARQL queries.}
To be included, each object must provide:
(1)~a digital image (\texttt{P18}), 
(2)~temporal metadata about its creation (e.g., \texttt{P571}),\footnote{If \enquote{inception} is not available, we use the properties \enquote{start time} (\texttt{P580}), \enquote{end time} (\texttt{P582}), and \enquote{publication date} (\texttt{P577}) to obtain time-related creation information.} and
(3)~an object-type label (\texttt{P31}).
Creator information (\texttt{P170}) is retained when available and utilized for diversity-aware sampling; records without known creators are not excluded.
The temporal information attached to each record is then normalized into an interval.
If Wikidata specifies a precise creation year, this interval collapses to a single point, such that the start and end years coincide; if, by contrast, the record provides a range, we retain the earliest and latest plausible years as the interval boundaries $y_A^{\mathrm{start}}$ and $y_A^{\mathrm{end}}$.\footnote{When multiple years or ranges are given, we define the interval by taking the earliest and latest plausible years across all entries.}
For vague date qualifiers such as \enquote{circa} (\texttt{Q5727902}), we expand the interval using a rule-based scheme with narrower uncertainty for more recent dates~\cite{riecke2009}; e.g., \enquote{circa 1400} becomes $[1380,1420]$ and \enquote{circa 2000} becomes $[1995,2005]$. 
For model fitting, each interval is converted into a scalar midpoint year $y_A$, as described in \cref{sec:methodology}.
%
To avoid sparsely represented periods, we restrict the experiments to $1400 < y_A \leq 2000$, yielding \num{12}~50-year strata from 1401--1450 to 1951--2000. 
Within each stratum, records are deduplicated and shuffled, and \num{500}~records are retained.
To increase the diversity of the sample, we preferentially select records such that known creators are not repeated within the same temporal stratum.
We restrict records to the four most frequent object-types: \enquote{painting,} \enquote{print,} \enquote{drawing,} and \enquote{sculpture.}
More fine-grained labels are mapped to these categories where appropriate (for instance, \enquote{watercolor painting} to \enquote{painting}).
Each stratum contains \num{125}~records per type, yielding \num{1500} per type overall. 
This balancing reduces the influence of the class-count differences present in the unfiltered Wikidata population on the temporal and object-type analyses, although it does not eliminate representation or imaging differences across media. 
Each temporal stratum is then split into training, validation, and test partitions; \SI{10}{\percent} of the sampled records are assigned to the validation set and \SI{10}{\percent} to the test set, while the remaining \SI{80}{\percent} are used for training.
Across the full temporal range, this results in \num{4800}~training, \num{600}~validation, and \num{600}~test records, yielding a total sample size of \num{6000} records.

\section{Experiments}
\label{sec:experiments}

In this section, we assess whether pretrained image representations contain sufficient temporal information to estimate the creation dates of artworks.
The question has two aspects.
On the one hand, it concerns predictive performance: can visual embeddings support the estimation of historical dates?
On the other hand, it concerns the structure of the temporal signal itself: what kinds of historical regularities become available to a model through digitized images and their metadata?
We assess the proposed approach along three dimensions:
(1)~the accuracy of creation-date estimates,
(2)~the reliability of uncertainty-aware prediction intervals, and
(3)~the variation of performance across historical periods and object types.
We describe the experimental setup (\cref{sec:experimental-setup}) and the results obtained through both quantitative (\cref{sec:experiments-quantitative}) and qualitative analyses (\cref{sec:experiments-qualitative}).

\subsection{Experimental Setup}
\label{sec:experimental-setup}

The main estimator is a quantile-regression model trained on image embeddings.
We estimate the predictive quantiles $\mathcal{Q}=\{0.10,0.50,0.90\}$, using the median $\hat{q}_{A,0.50}$ as the central creation-date estimate $\hat{y}_A$.
The lower and upper quantiles $(0.10,0.90)$ define the \SI{80}{\percent} prediction interval $\hat{D}_{A}^{(0.10,0.90)}$.

\paragraph{Baselines}
\label{sec:baseline}

We include two non-parametric baselines:
(1)~an empirical quantile baseline, which ignores the image and, for every test example, predicts the corresponding quantiles of the training-set distribution.
It therefore captures the temporal prior of the dataset, i.e., the information available from the marginal distribution of the creation dates alone; 
(2)~a $k$-nearest-neighbor baseline in the frozen visual embedding space.
For each test record, we retrieve the \num{50}~nearest training records based on cosine distance and estimate the target quantiles from the empirical distribution of their creation dates; we set $k=50$ as a compromise between locality and reliable tail-quantile estimation, yielding approximately five neighbors in each of the $0.10$ and $0.90$ tails.
This baseline uses the same frozen visual representations as the learned regressors but does not learn an explicit supervised mapping from embedding coordinates to creation year. 

\paragraph{Evaluation Metrics}
\label{sec:evaluation-metrics}

Consistent with prior work, point-estimation performance is evaluated using \ac{MAE} and \ac{RMSE}~\cite{heginbotham2018, strezoski2018}.
We treat \ac{MAE} as the primary point-estimation metric because it is directly interpretable in years and, unlike \ac{RMSE}, is not dominated by a small number of large temporal errors; \ac{RMSE} is nevertheless reported as a measure, since large deviations are precisely the cases in which a dating model can become historically misleading.
Uncertainty estimates are evaluated through empirical interval coverage and mean interval width.
For a predicted interval, 
coverage is the proportion of test examples for which the reference midpoint year $y_A$ falls inside the interval.
Coverage and width therefore have to be read together: a useful temporal interval should contain the reference years reliably; at the same time, it should not achieve this reliability simply by expanding until it becomes historically uninformative.
We additionally report the mean Winkler interval score, $\overline{W}_{\alpha}$~\cite{gneiting2007,winkler1972}. 
The score combines interval width with an explicit penalty when the target falls outside the predicted interval; lower values indicate better performance. 
Here, $\alpha=0.20$ denotes the nominal non-coverage rate of the reported \SI{80}{\percent} prediction interval and is not a tuned hyperparameter. 
Finally, we compare the predicted interval $\hat{D}_A$ with the original catalog interval $D_A$.
The overlap metric $\cap$~$D_A$ denotes the percentage of test examples for which $\hat{D}_{A}$ has a non-empty intersection with $D_A$, whereas the containment metric $\supseteq$~$D_A$ denotes the percentage of examples for which $\hat{D}_{A}$ fully contains $D_A$; the latter specifically measures whether the predicted uncertainty range fully covers the uncertainty represented by the catalog metadata.
We do not report length-based Jaccard scores because zero-width catalog intervals yield zero overlap even when the predicted interval contains the exact year. 

\paragraph{Implementation Details}
\label{sec:implementation-details}

The evaluation is conducted using the following \acp{VLM}:
(1)~CLIP ViT-B/16~\cite{clip};
(2)~EVA02-CLIP-B/16~\cite{evaclip}; and
(3)~SigLIP2-B/16~\cite{siglip2}.
To assess whether the observed effects are specific to vision--language pretraining, we additionally include two vision-only self-supervised baselines with DINOv2-B/14~\cite{dinov2} and I-JEPA ViT-H/14~\cite{ijepa}.
For each backbone, we evaluate two prediction configurations: a standard quantile-regression setting and a conformalized variant.
Both configurations use LightGBM~\cite{lightgbm} as the regression backend.\footnote{We also evaluated gradient boosting as implemented in scikit-learn \cite{scikit-learn}, but observed only small and non-systematic effects.}
For each target quantile, an independent regressor is trained with \num{2000} estimators, a learning rate of \num{0.05}, and a maximum tree depth of \num{5}.\footnote{Multiple parameter configurations were tested; the best-performing configuration was then used for further evaluation.}
Before regression, all embeddings are reduced to \num{128}~principal components using \ac{PCA}.
The quantile-regression models are fitted only on the training subset; the validation subset is then used to estimate the split-conformal correction, which is applied to the lower and upper quantile predictions.

\subsection{Quantitative Results}
\label{sec:experiments-quantitative}

\begin{table}[t!]
    \caption{
        Test-set performance of quantile-regression estimators and $k$-nearest-neighbor baselines. 
        Point-estimation metrics are evaluated against the midpoint year $y_A$ and reported as \ac{MAE} and \ac{RMSE} in years.
        Interval performance is reported for the \SI{80}{\percent} prediction interval $\hat{D}_{A}^{(0.10,0.90)}$.
        Metrics are defined in \cref{sec:evaluation-metrics}.
        Rows marked with \checkmark{} apply split-conformal calibration.
        Best values are shown in bold.
    }
    \label{tab:quantitative-results}

    \scriptsize
    \begin{tabularx}{\linewidth}{@{}X p{1.5cm} p{0.8cm} RR RRRRR@{}}
        \toprule
        \multirow{2}{*}{Backbone} & \multirow{2}{*}{Estimator} & \multirow{2}{*}{Cal.}
        & \multicolumn{2}{c}{Point Estimate}
        & \multicolumn{5}{c}{Prediction Interval} \\
        \cmidrule{4-5}
        \cmidrule(l){6-10}
        & &
        & MAE & RMSE
        & Cov. & Width & $\overline{W}_{\alpha}$
        & $\cap$ $D_A$ & $\supseteq$ $D_A$ \\
    
        \midrule
        \multirow{2}{*}{--}
        & Prior &              & 143.6 & 165.0 & 82.5 & 448.0 & 513.9 & 85.2 & 79.2 \\
        & Prior & \checkmark{} & 143.5 & 165.0 & 84.3 & 469.3 & 517.4 & 86.7 & 81.7 \\
    
        \midrule
        \multirow{4}{*}{CLIP}
        & QR  &              & 66.1  & 89.3  & 63.7 & \textbf{178.2} & 329.5 & 69.0 & 59.5 \\
        & QR  & \checkmark{} & \textbf{65.7} & \textbf{88.5} & 80.5 & 234.5 & \textbf{308.9} & 82.5 & 75.7 \\
        \cmidrule{2-10}
        & kNN &              & 70.4  & 100.3 & 85.0 & 255.8 & 313.4 & \textbf{87.7} & 79.8 \\
        & kNN & \checkmark{} & 71.2  & 100.8 & 85.5 & 259.1 & 314.7 & \textbf{87.7} & \textbf{80.5} \\
    
        \midrule
        \multirow{4}{*}{EVA-CLIP}
        & QR  &              & 70.0  & 94.8  & 61.2 & 186.9 & 348.2 & 65.3 & 55.2 \\
        & QR  & \checkmark{} & 70.8  & 94.5  & \textbf{80.2} & 254.1 & 330.7 & 82.2 & 76.3 \\
        \cmidrule{2-10}
        & kNN &              & 75.8  & 109.5 & 81.2 & 249.3 & 331.5 & 84.2 & 75.7 \\
        & kNN & \checkmark{} & 76.7  & 109.6 & 82.5 & 254.1 & 332.7 & 85.3 & 76.8 \\
    
        \midrule
        \multirow{4}{*}{SigLIP2}
        & QR  &              & 66.2  & 89.6  & 58.0 & 184.0 & 341.0 & 62.5 & 52.3 \\
        & QR  & \checkmark{} & 67.7  & 89.4  & 82.0 & 250.5 & 319.7 & 83.3 & 77.7 \\
        \cmidrule{2-10}
        & kNN &              & 74.6  & 107.1 & 83.5 & 264.2 & 324.3 & 85.7 & 78.8 \\
        & kNN & \checkmark{} & 74.9  & 105.7 & 84.0 & 269.3 & 327.9 & 86.0 & 79.3 \\
    
        \midrule
        \multirow{4}{*}{DINOv2}
        & QR  &              & 86.2  & 115.0 & 60.0 & 235.2 & 433.3 & 62.5 & 54.8 \\
        & QR  & \checkmark{} & 86.5  & 114.1 & 79.2 & 316.7 & 394.6 & 81.3 & 76.7 \\
        \cmidrule{2-10}
        & kNN &              & 96.8  & 131.8 & 76.3 & 290.0 & 412.7 & 79.0 & 72.0 \\
        & kNN & \checkmark{} & 97.9  & 132.2 & 79.2 & 303.6 & 413.6 & 81.8 & 74.8 \\
    
        \midrule
        \multirow{4}{*}{I-JEPA}
        & QR  &              & 98.2  & 123.5 & 58.2 & 265.3 & 457.6 & 61.5 & 54.3 \\
        & QR  & \checkmark{} & 101.7 & 127.1 & \textbf{79.8} & 353.2 & 434.3 & 82.2 & 77.5 \\
        \cmidrule{2-10}
        & kNN &              & 103.7 & 134.8 & 77.2 & 329.0 & 416.2 & 80.8 & 73.2 \\
        & kNN & \checkmark{} & 105.8 & 136.1 & 78.0 & 338.4 & 419.0 & 81.2 & 73.7 \\
        \bottomrule
    \end{tabularx}
\end{table}

\paragraph{Backbone Comparison}

The results indicate that, while pretrained image representations do encode usable temporal information, this information is distributed unevenly across embedding spaces and calibration settings (\cref{tab:quantitative-results}).
No single configuration dominates all metrics.
CLIP with split-conformal calibration achieves the lowest \ac{MAE} and \ac{RMSE}, while SigLIP2 with split-conformal calibration achieves the highest midpoint coverage and the strongest agreement with the catalog intervals.
Median-year errors remain comparable across CLIP, EVA-CLIP, and SigLIP2, with calibrated \ac{MAE} values between \num{65.7} and \num{70.8}~years.
The performance gap between \acp{VLM} and purely visual self-supervised representations suggests that image--text pretraining is advantageous for temporal estimation.
The relevant signal is therefore unlikely to be purely formal or low-level; instead, \acp{VLM} appear to encode visual-semantic regularities associated with historically shaped features, as has been observed to a lesser degree in prior work~\cite{kim2025,li2018,strafforello2025}.
The $k$-nearest-neighbor baseline supports this interpretation: its improvement over the empirical prior shows that local neighborhoods exhibit chronological structure.
However, its lower accuracy compared to the learned quantile regression models suggests that temporal information cannot be fully captured by nearest-neighbor similarity alone.

\paragraph{Interval Calibration}

Split-conformal calibration improves the empirical coverage of the $\hat{D}_{A}^{(0.10,0.90)}$ intervals, bringing the learned models close to or slightly above the nominal \SI{80}{\percent} level; as expected, this increase is accompanied by wider prediction intervals (\cref{tab:quantitative-results}).
The Winkler scores indicate that conformal calibration enhances the overall trade-off between informativeness and reliable coverage.
The empirical prior attains nominal coverage, but does so only by producing very broad intervals and much weaker point estimates.
In contrast, calibrated \acp{VLM} provide a more favorable balance: after conformal calibration, their predicted intervals overlap the original catalog intervals for approximately \SI{82}{\percent} to \SI{83}{\percent} of test examples, compared with \SI{63}{\percent} to \SI{69}{\percent} before calibration.

\paragraph{Temporal Error Structure}

As shown in \cref{tab:siglip2-century-analysis} for SigLIP2 with split-conformal calibration, the mean signed error changes direction across the chronological range: objects from the fifteenth to seventeenth centuries are, on average, dated too late, whereas objects from the eighteenth to twentieth centuries are dated too early.
Predictions for uncertain cases therefore tend to be drawn toward the most statistically stable regions of the learned temporal distribution.
The quantile-regression objective may further reinforce this tendency.
In other words, when visual evidence is not period-specific, or when an object does not conform to a stabilized art-historical vocabulary, the loss can be minimized through predictions that remain close to the center of the distribution, rather than by preserving historically specific deviations from it.

\begin{table}[t!]
    \caption{
        Test-set performance by century of the SigLIP2-based quantile-regression estimator with split-conformal calibration.
        Point-estimation metrics are computed against the midpoint year $y_A$ and reported as \ac{MAE}, \ac{RMSE}, and signed bias $\bar{e}$ in years.
        Interval performance is reported for the \SI{80}{\percent} prediction interval $\hat{D}_{A}^{(0.10,0.90)}$.
        Metrics are defined in \cref{sec:evaluation-metrics}.
    }
    \label{tab:siglip2-century-analysis}
    
    \scriptsize
    \begin{tabularx}{\linewidth}{@{}X RRR RRRRR@{}}
        \toprule
        \multirow{2}{*}{Century}
        & \multicolumn{3}{c}{Point Estimate}
        & \multicolumn{5}{c}{Prediction Interval} \\
        \cmidrule{2-4}
        \cmidrule(l){5-9}
        & MAE & RMSE & Bias ($\bar{e}$)
        & Cov. & Width & $\overline{W}_{\alpha}$
        & $\cap$ $D_A$ & $\supseteq$ $D_A$ \\
    
        \midrule
        1401--1500 & 90.7 & 115.3 & +87.4 & 60.0 & 304.3 & 483.4 & 64.3 & 50.0 \\
        1501--1600 & 76.2 & 101.2 & +66.5 & 78.0 & 262.6 & 351.7 & 80.0 & 71.0 \\
        1601--1700 & 56.1 & 74.1  & +32.2 & 90.8 & 249.3 & 272.9 & 92.7 & 90.8 \\
        1701--1800 & 55.0 & 69.1  & -32.5 & 95.4 & 230.4 & 241.3 & 96.3 & 91.7 \\
        1801--1900 & 58.2 & 74.3  & -43.7 & 90.2 & 236.4 & 249.0 & 90.2 & 87.3 \\
        1901--2000 & 74.9 & 97.8  & -69.0 & 73.1 & 236.1 & 346.9 & 73.1 & 68.3 \\
        \bottomrule
    \end{tabularx}
\end{table}

\paragraph{Object-type Effects}

Paintings and prints yield the most accurate point estimates, with \acp{MAE} of \num{55.9} and \num{59.6}~years, respectively, and both meet or exceed the nominal \SI{80}{\percent} interval coverage (\cref{tab:siglip2-object-type-analysis}).
Drawings show a moderately higher \ac{MAE}, but achieve the highest interval coverage, as well as the strongest catalog-interval overlap and containment.
The model therefore often captures the broader chronological uncertainty encoded in the catalog metadata, even when the point estimates are less accurate than those for paintings and prints.
Sculptures, by contrast, are the clearest outlier: they produce the highest \ac{MAE} and \ac{RMSE}, the lowest coverage, and the widest prediction intervals.
The model therefore localizes their creation dates less accurately while producing broader uncertainty estimates that remain less reliably calibrated.
Because the corpus is balanced by object type within each temporal stratum, this performance gap cannot be attributed to the large class-count imbalance of the source population. 
Several non-exclusive explanations remain plausible, however, and the present study design cannot distinguish among them. 
In particular, a single 2D image provides only a partial representation of a 3D object, with photographs of sculptures varying substantially in viewpoint, scale, and lighting. 
We therefore interpret these results as evidence of an object-type performance gap rather than evidence for any specific causal mechanism. 

\begin{table}[t!]
    \caption{
        Test-set performance by object type of the SigLIP2-based quantile-regression estimator with split-conformal calibration.
        Point-estimation metrics are computed against the midpoint year $y_A$ and reported as \ac{MAE}, \ac{RMSE}, and signed bias $\bar{e}$ in years.
        Interval performance is reported for the \SI{80}{\percent} prediction interval $\hat{D}_{A}^{(0.10,0.90)}$.
        Metrics are defined in \cref{sec:evaluation-metrics}.
    }
    \label{tab:siglip2-object-type-analysis}
    
    \scriptsize
    \begin{tabularx}{\linewidth}{@{}X RRR RRRRR@{}}
        \toprule
        \multirow{2}{*}{Object type}
        & \multicolumn{3}{c}{Point Estimate}
        & \multicolumn{5}{c}{Prediction Interval} \\
        \cmidrule{2-4}
        \cmidrule(l){5-9}
        & MAE & RMSE & Bias ($\bar{e}$)
        & Cov. & Width & $\overline{W}_{\alpha}$
        & $\cap$ $D_A$ & $\supseteq$ $D_A$ \\
    
        \midrule
        Painting  & 55.9 & 73.1  & -9.8  & 82.7 & 234.2 & 291.0 & 84.0 & 78.7 \\
        Sculpture & 91.6 & 115.9 & +18.8 & 73.3 & 290.7 & 405.3 & 74.7 & 70.7 \\
        Print     & 59.6 & 80.3  & -5.0  & 84.7 & 229.0 & 288.8 & 86.7 & 81.3 \\
        Drawing   & 63.9 & 82.0  & -2.2  & 87.3 & 248.0 & 293.6 & 88.0 & 80.0 \\
        \bottomrule
    \end{tabularx}
\end{table}

\subsection{Qualitative Results}
\label{sec:experiments-qualitative}

While the quantitative results demonstrate that visual-semantic regularities can support artwork dating, they do not explain why the resulting temporal signal may be misleading.
Therefore, we complement the quantitative evaluation with a qualitative analysis of selected predictions obtained by SigLIP2 under split-conformal calibration.
This analysis revisits the question raised in the introduction: do pretrained visual representations encode historical time directly, or do they instead organize artworks within a perceptual topology shaped by factors such as preservation, cataloging, and digitization?
We investigate this question through selected examples, grouped by the type of bias observable in each case~\cite{friedman1996,suresh2021}.
These examples are intended to be diagnostic rather than representative: they are neither randomly sampled nor used to estimate prevalence; accordingly, they support hypotheses about potential biases but do not establish how frequently those biases occur in the corpus. 

\paragraph{Representation Bias}

The low-error cases mostly share a high degree of visual specificity; in these cases, the predicted median year $\hat{y}_A$ falls close to the catalog date, sometimes differing by only a few years (\crefrange{fig:masolino-annunciation}{fig:marquet-bougie-port-of-algeria}).
Two forms of representation bias may contribute to this pattern: 
first, the pretrained encoder may represent frequently encountered canonical styles and subjects more effectively as a consequence of its pretraining distribution; 
second, the Wikidata-derived regression corpus reflects the selective processes through which institutions and digitization projects preserve, catalog, and make artworks available as images. 
Both factors can render some visual-historical vocabularies more learnable than others. 
As the present experiment cannot disentangle these contributions, we employ the term \textit{representation bias} to refer to their combined effect rather than attributing the observed pattern to the pretraining distribution alone. 

\paragraph{Measurement Bias}

Although the supervised target is the creation date, the visual evidence available to the model may correspond to style, iconography, or other forms of historically situated visual language.
This can be understood as a form of measurement bias: while the target variable records one temporal attribute of the object, the image may emphasize other temporal associations more effectively.
A particularly instructive example is de Coxie's \textit{Design for the Front of a Pulpit} (1961; \cref{fig:de-coxie-anoniem}), which results in a significant prediction error.
Although cataloged as a twentieth-century drawing, the work contains religious subject matter and ornamental vocabulary associated with an earlier period; it appears that the model is estimating the date of the visual language evoked by the object rather than its production date.
This error illustrates the temporal entanglement at issue: the model responds to historically meaningful visual cues, whereas the supervised task requires these cues to be mapped onto a single creation date.
Kim \etal~\cite{kim2025} similarly find that semantic and iconographic information can be highly salient for periodization in art-historical images; our example shows how such cues may become misleading when the temporality of a work's visual language diverges from its production date. 

Other errors are better understood as failures of temporal specificity.
Snyders's painting, for example, is neither a revival of an iconographic theme nor a modern copy; it is a seventeenth-century still life (\cref{fig:snyders-still-life}).
Yet the model predicts it as being more than a century later.
This suggests that some object- and genre-level features remain visually stable across long historical periods: while the model recognizes the work as a still life, it fails to locate it securely within a Flemish Baroque chronology.


\definecolor{linegray}{HTML}{404040}


\pgfdeclarelayer{predictionlinks}
\pgfdeclarelayer{foreground}
\pgfsetlayers{main,predictionlinks,foreground}


\tikzset{
  timeline axis/.style={
    draw=linegray,
    line width=0.5pt,
    -{Latex[length=2mm]}
  },
  timeline tick/.style={
    draw=linegray!55,
    line width=0.5pt
  },
  actual mark/.style={
    circle,
    fill=#1,
    draw=#1,
    inner sep=0pt,
    minimum size=4pt
  },
  predicted mark/.style={
    circle,
    fill=white,
    draw=#1,
    line width=0.5pt,
    inner sep=0pt,
    minimum size=4pt
  },
  artwork frame/.style={
    draw=#1,
    line width=1pt,
    inner sep=0pt,
    outer sep=0pt
  },
  timeline text/.style={
    font=\tiny,
    align=center,
    text=linegray,
    fill=white,
    inner sep=1pt
  },
  timeline interval/.style={
    line width=0.5pt,
    line cap=round,
    opacity=0.75
  },
  timeline interval stem/.style={
    line width=0.5pt,
    densely dashed,
    opacity=0.75
  },
  ground truth stem/.style={
    draw=#1,
    line width=0.5pt,
    preaction={
      draw=white,
      line width=2.5pt,
      opacity=1
    }
  }
}


\newcommand{\TimelineYearX}[1]{%
  {((#1 - 1400) / 600) * 13.2}%
}


\newcommand{\TimelineGroundTruthStemWithOwnLink}[4]{%

  \draw[
    white,
    line width=2.6pt,
    opacity=1
  ]
  (#2) -- ($(#2)!0.94!(#4)$);

  \draw[
    white,
    line width=2.6pt,
    opacity=1
  ]
  ($(#2)!1.06!(#4)$) -- (#3);

  \draw[
    #1,
    line width=0.65pt
  ]
  (#2) -- (#3);
}


\newcommand{\TimelineImageXYNoStems}[9]{%

  \refstepcounter{subfigure}%
  \label{#8}%

  \coordinate (actual-#2) at (\TimelineYearX{#3},0);
  \coordinate (image-#2)  at (\TimelineYearX{#3},#5);
  \coordinate (actual-lane-#2) at (\TimelineYearX{#3},#9);

  \begin{pgfonlayer}{foreground}
    \draw[ground truth stem=#6] (actual-#2) -- (image-#2);

    \node[actual mark=#6] at (actual-#2) {};

    \node[
      artwork frame=#6,
      anchor=center
    ] (img-#2) at (image-#2)
    {\href{https://www.wikidata.org/wiki/#2}{%
      \includegraphics[height=1.55cm]{#1}%
    }};

    \node[
      font=\scriptsize,
      anchor=north,
      yshift=-2pt
    ] at (img-#2.south) {(\thesubfigure)};
  \end{pgfonlayer}
}


\newcommand{\TimelineImageXYWithStems}[9]{%

  \refstepcounter{subfigure}%
  \label{#8}%

  \coordinate (actual-#2) at (\TimelineYearX{#3},0);
  \coordinate (pred-#2)   at (\TimelineYearX{#4},0);
  \coordinate (image-#2)  at (\TimelineYearX{#3},#5);
  \coordinate (actual-lane-#2) at (\TimelineYearX{#3},#9);

  \begin{pgfonlayer}{foreground}
    \TimelineGroundTruthStemWithOwnLink
      {#6}
      {actual-#2}
      {image-#2}
      {actual-lane-#2}

    \node[actual mark=#6] at (actual-#2) {};
    \node[predicted mark=#6] at (pred-#2) {};

    \node[
      artwork frame=#6,
      anchor=center
    ] (img-#2) at (image-#2)
    {\href{https://www.wikidata.org/wiki/#2}{%
      \includegraphics[height=1.55cm]{#1}%
    }};

    \node[
      font=\scriptsize,
      anchor=south,
      yshift=2pt
    ] at (img-#2.north) {(\thesubfigure)};
  \end{pgfonlayer}

  \begin{pgfonlayer}{predictionlinks}
    \draw[
      #6,
      timeline interval,
      densely dotted
    ] ($(img-#2.south)+(0,-0.75pt)$) -- (pred-#2);
  \end{pgfonlayer}
}


\begin{figure}[t]
  \centering
  \scriptsize

  \stepcounter{figure}
  \setcounter{subfigure}{0}

  \resizebox{\linewidth}{!}{%
  \begin{tikzpicture}[x=1cm,y=1cm]
    \draw[timeline axis] (-0.6,0) -- (14,0);

    \foreach \yr in {1400,1500,1600,1700,1800,1900,2000}{
      \draw[timeline tick]
        (\TimelineYearX{\yr},0.09) --
        (\TimelineYearX{\yr},-0.09);

      \node[font=\tiny, below=5pt, text=linegray]
        at (\TimelineYearX{\yr},0) {\yr};
    }


    \TimelineImageXYWithStems
      {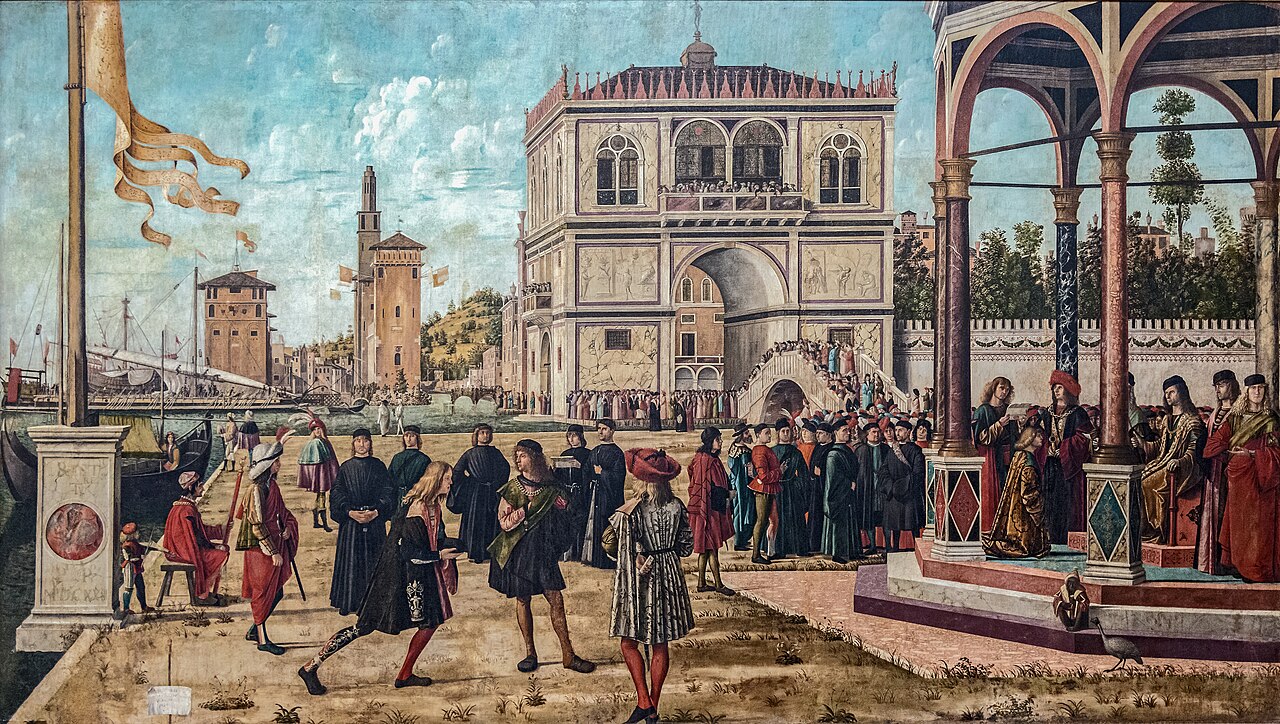}
      {Q3937363}
      {1498}
      {1773.8}
      {1.6}
      {red!75!black}
      {Carpaccio}
      {fig:carpaccio-accademia}
      {0.5}

    \TimelineImageXYWithStems
      {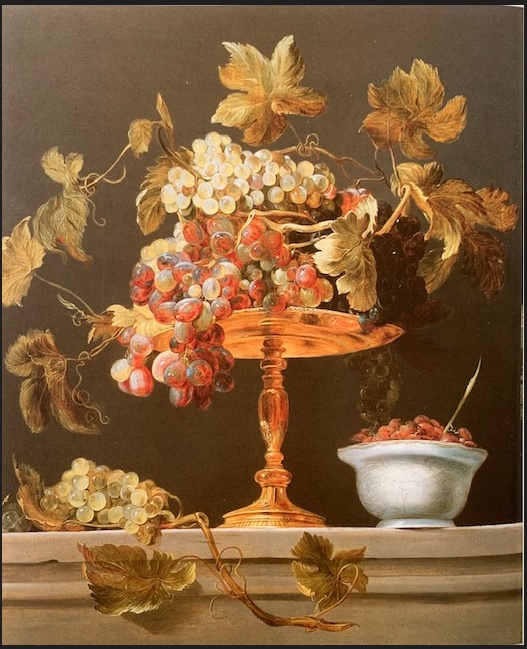}
      {Q113861785}
      {1650}
      {1784.3}
      {1.6}
      {red!75!black}
      {Snyders}
      {fig:snyders-still-life}
      {0.4}

    \TimelineImageXYWithStems
      {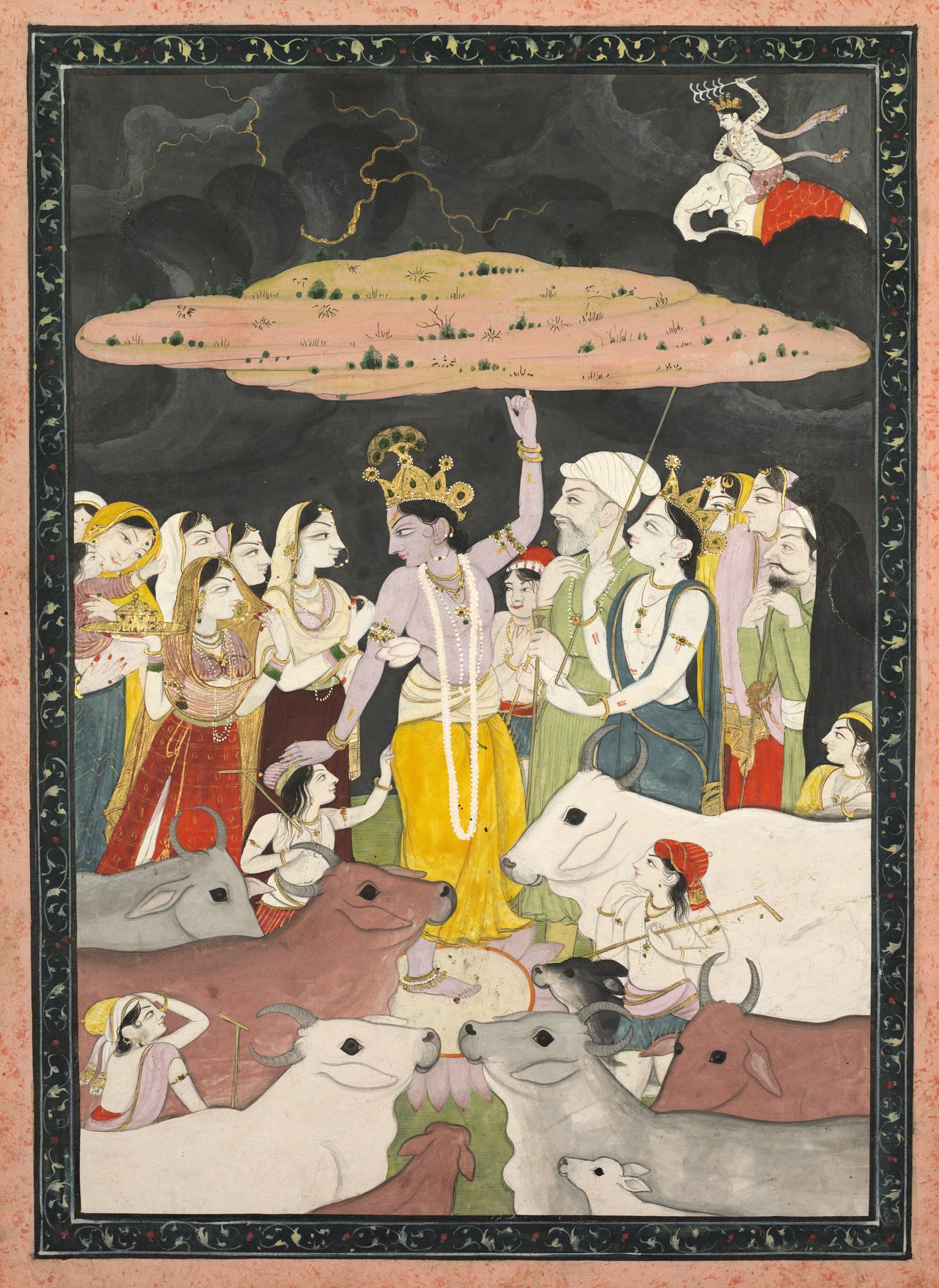}
      {Q60478335}
      {1785}
      {1636.2}
      {1.6}
      {red!75!black}
      {Unknown}
      {fig:unknown-krishna-lifting-mount-govardhan}
      {0.2}

    \TimelineImageXYWithStems
      {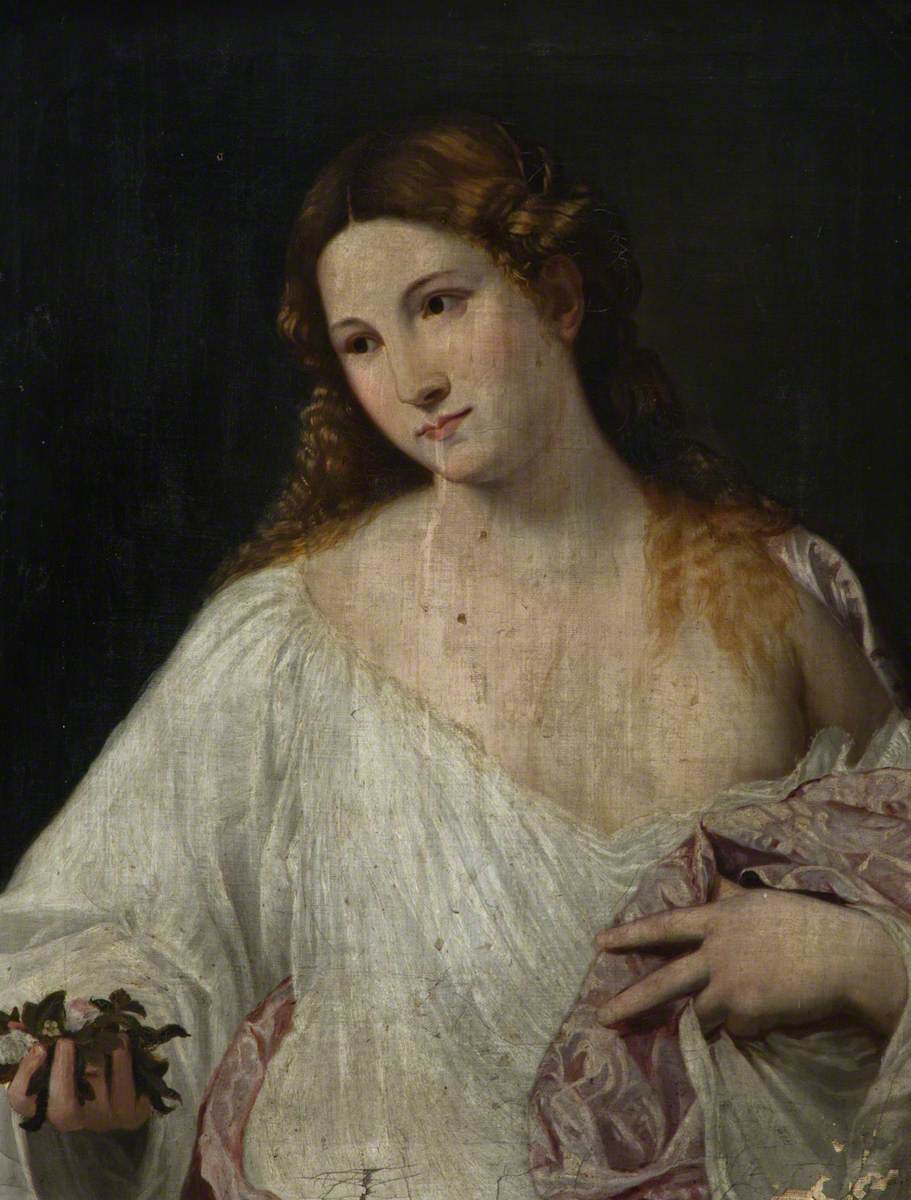}
      {Q120064383}
      {1890}
      {1664.9}
      {1.6}
      {red!75!black}
      {Fraser}
      {fig:fraser-flora}
      {0.3}

    \TimelineImageXYWithStems
      {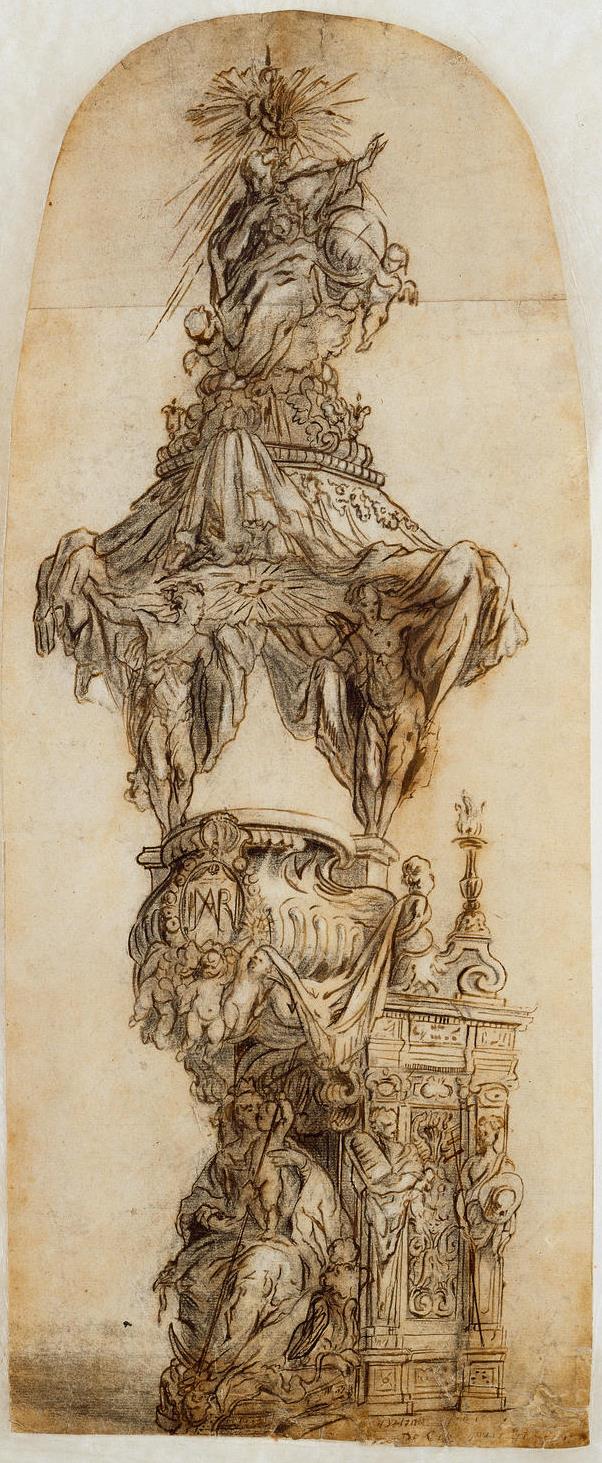}
      {Q59097398}
      {1961}
      {1545.4}
      {1.6}
      {red!75!black}
      {de Coxie}
      {fig:de-coxie-anoniem}
      {0.6}

    \TimelineImageXYNoStems
      {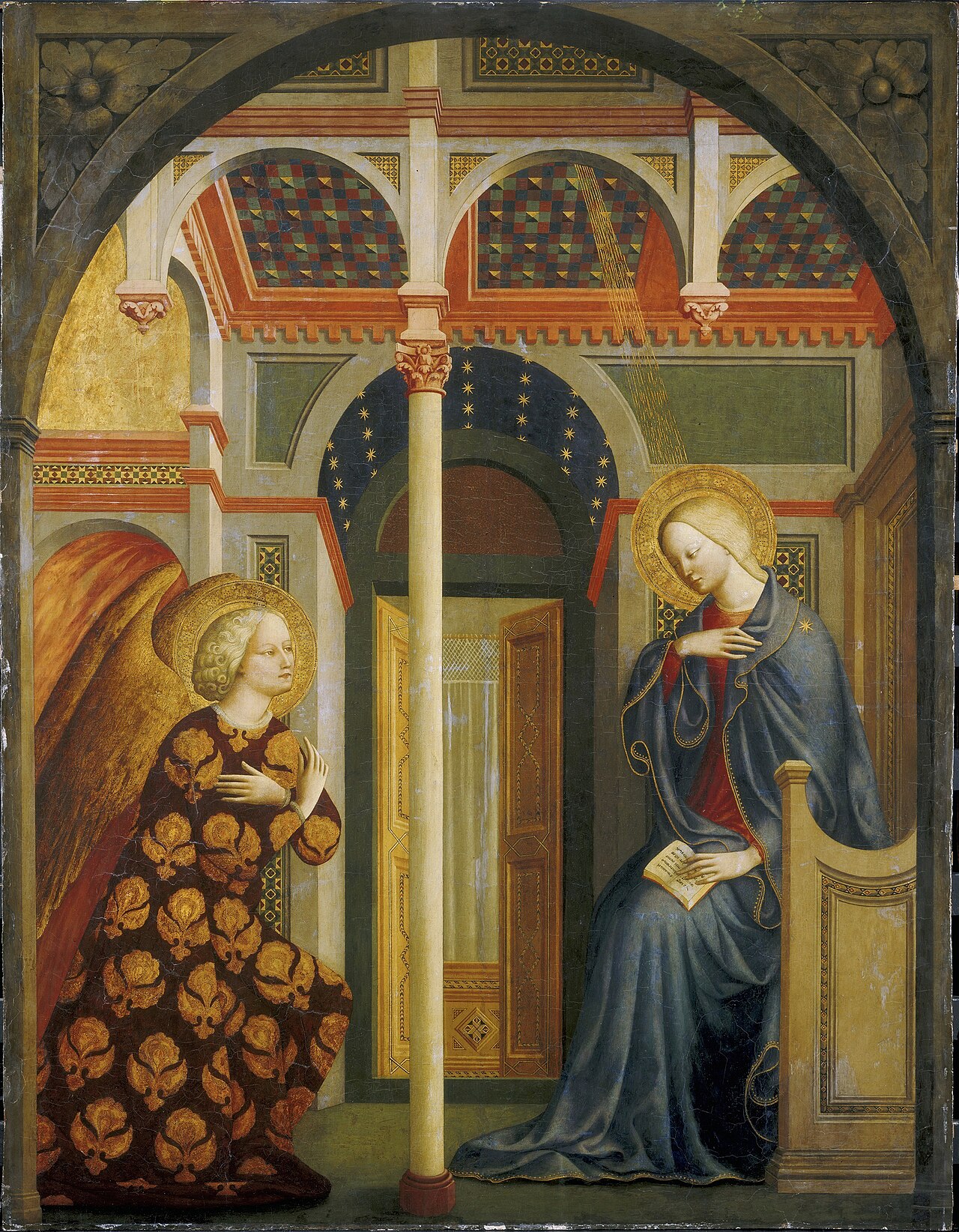}
      {Q3618180}
      {1423}
      {1430.9}
      {-1.6}
      {green!55!black}
      {Masolino}
      {fig:masolino-annunciation}
      {0}

    \TimelineImageXYNoStems
      {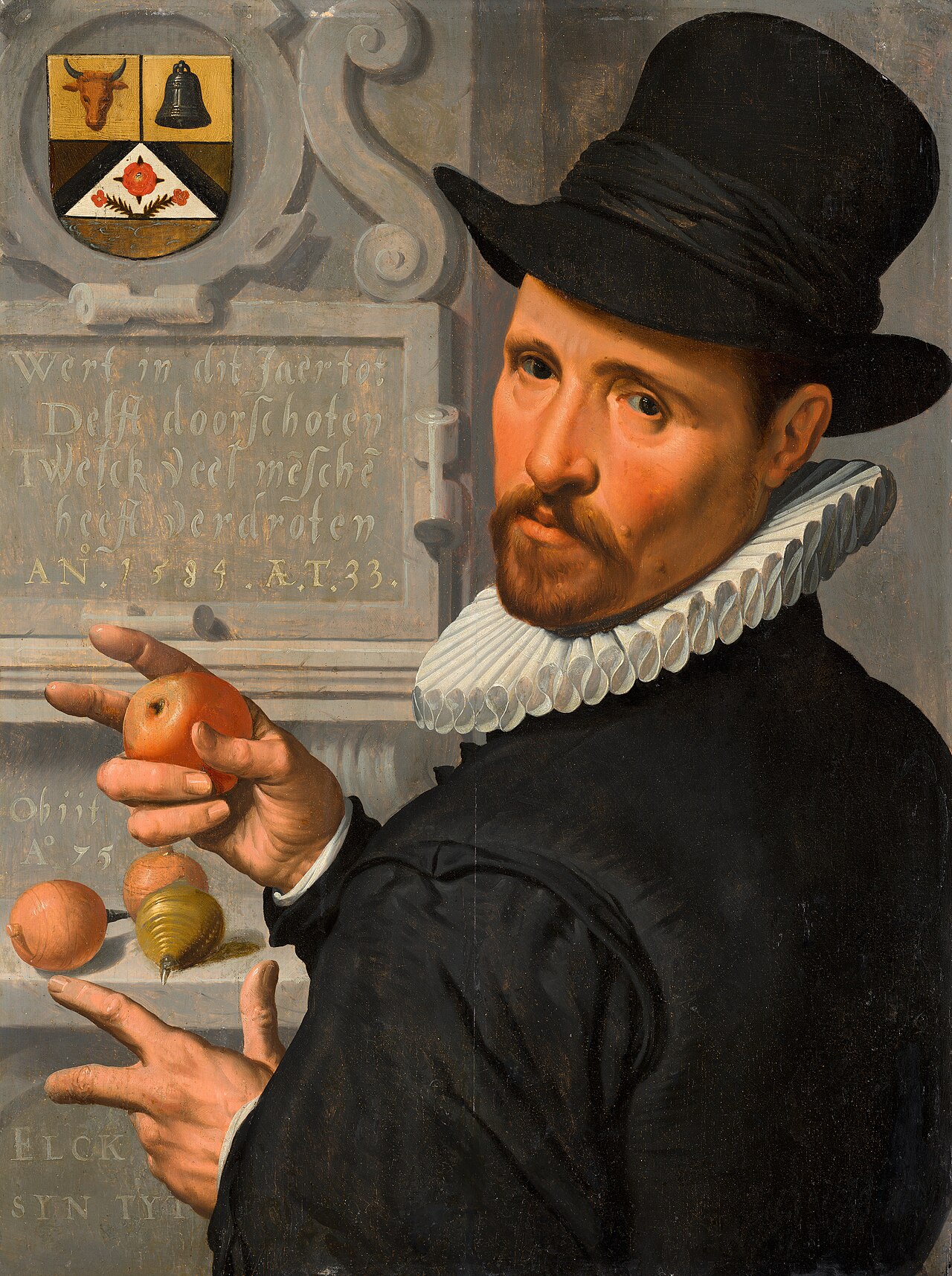}
      {Q17275707}
      {1584}
      {1579.4}
      {-1.6}
      {green!55!black}
      {Pietersz}
      {fig:pietersz-portrait-of-schellinger}
      {0}

    \TimelineImageXYNoStems
      {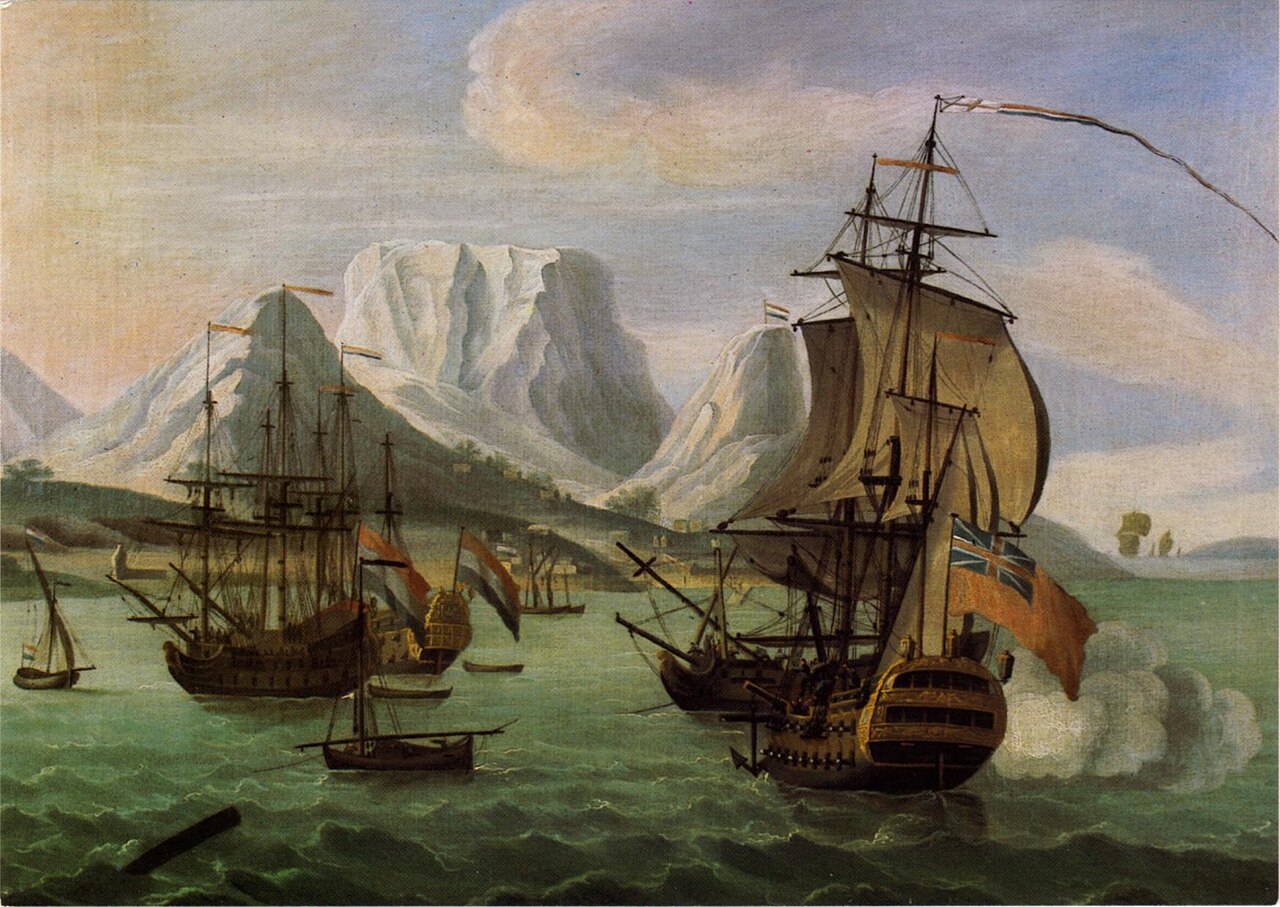}
      {Q55935695}
      {1730}
      {1733.2}
      {-1.6}
      {green!55!black}
      {Scott}
      {fig:scott-table-bay}
      {0}

    \TimelineImageXYNoStems
      {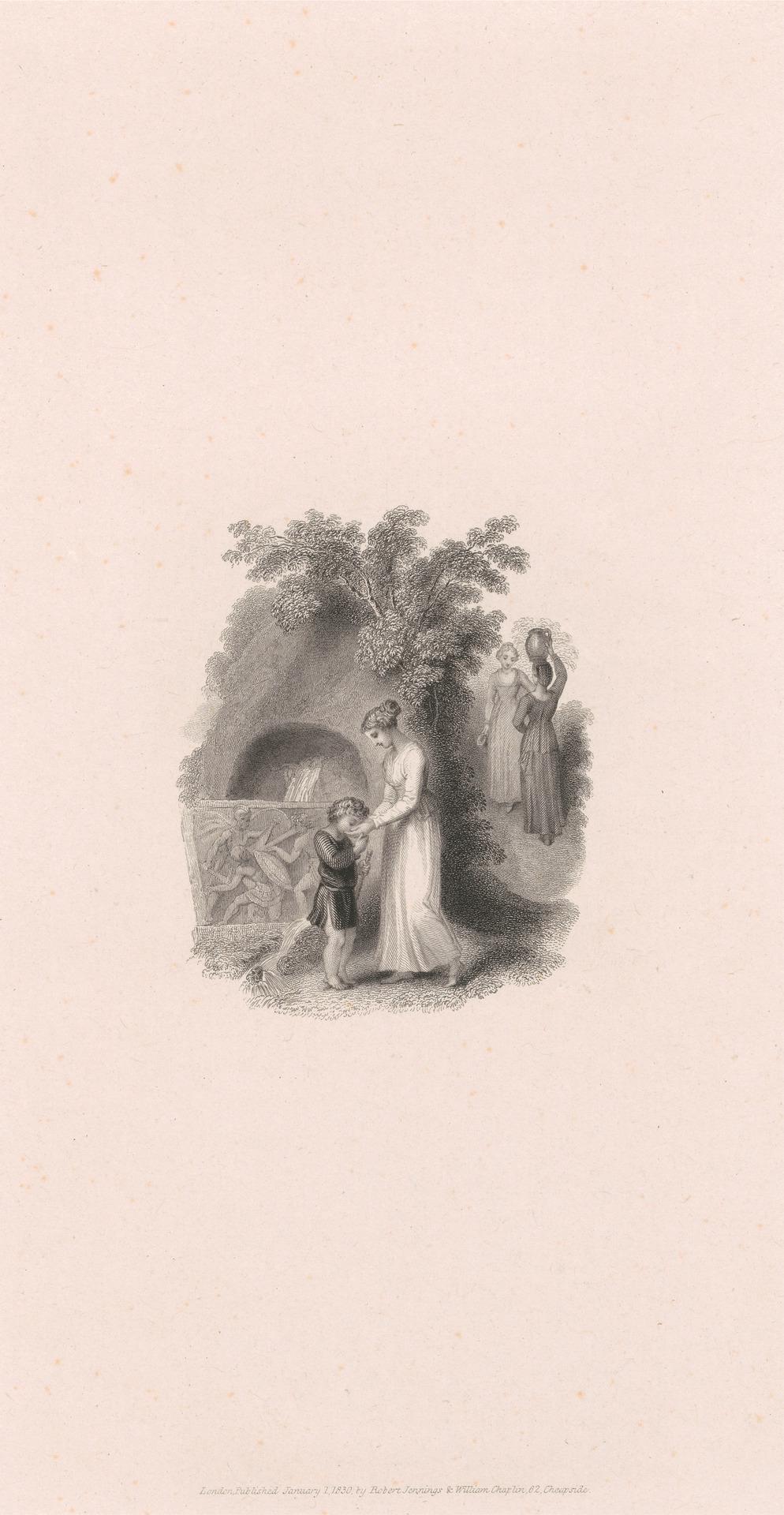}
      {Q110126199}
      {1830}
      {1828.2}
      {-1.6}
      {green!55!black}
      {Finden}
      {fig:finden-the-fountain}
      {0}

    \TimelineImageXYNoStems
      {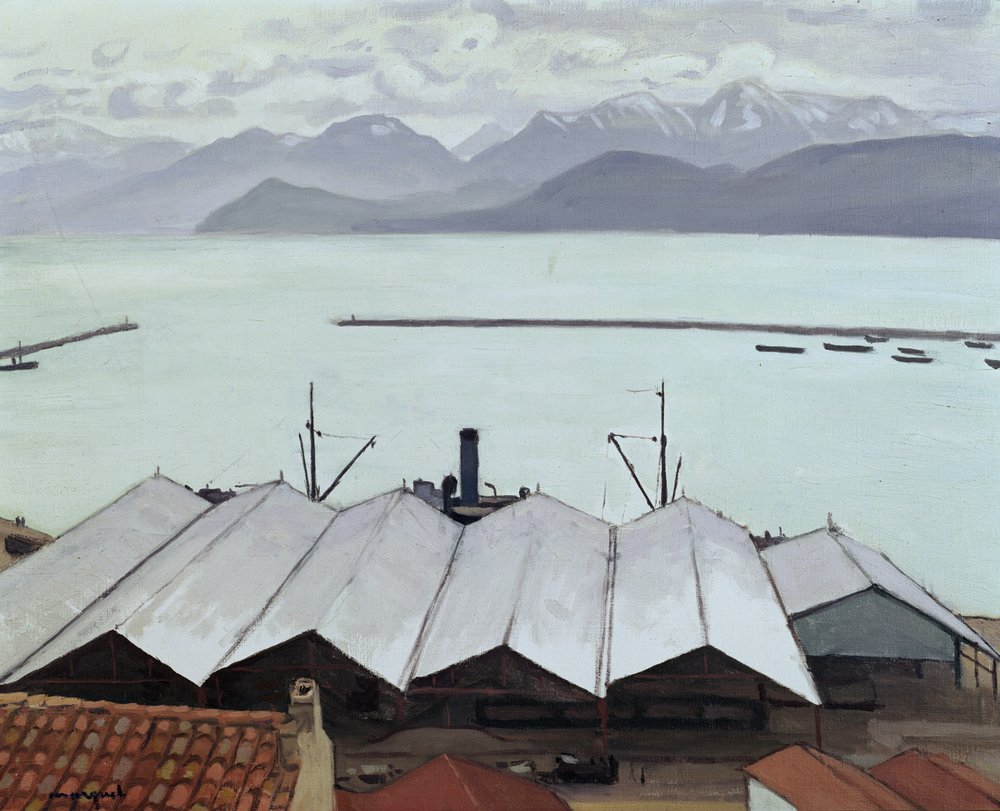}
      {Q114246369}
      {1926}
      {1923.3}
      {-1.6}
      {green!55!black}
      {Marquet}
      {fig:marquet-bougie-port-of-algeria}
      {0}
  \end{tikzpicture}%
  }

  \addtocounter{figure}{-1}

  \caption{
    Qualitative examples of SigLIP2 date predictions with split-conformal calibration.
    Filled markers indicate cataloged production dates and hollow markers model predictions.
    \textit{Green} denotes low-error cases, with absolute errors of up to \num{25}~years;
    \textit{red} denotes larger errors, with absolute errors of more than \num{50}~years.
  }
  \label{fig:qualitative-results}
\end{figure}

\paragraph{Aggregation Bias}

As illustrated by \cref{fig:unknown-krishna-lifting-mount-govardhan}, non-Western objects frequently appear to be positioned within the chronology of the dominant European art-historical canon rather than within their own regional and historical trajectories.\footnote{A systematic analysis would require more detailed geographical and cultural metadata in Wikidata. As such information is often missing or inconsistently encoded for non-Western artifacts, the necessary curation is beyond the scope of this paper.}
Ananthram \etal~\cite{ananthram2025} likewise document cultural performance gaps in \acp{VLM}. 
In our setting, this can be described as aggregation bias: a single temporal model is fitted across heterogeneous artistic traditions whose periodizations are not necessarily commensurable. 
The resulting errors suggest that the model may collapse these heterogeneous histories into the temporal structure most strongly represented in its learned distribution.
In such cases, non-dominant artistic traditions become legible to the model only insofar as they can be related to the visual and chronological patterns it represents most strongly.
Therefore, the issue is not only incorrect dating, but also the imposition of an inappropriate temporal frame, in which the historical development of one artistic tradition is interpreted through the chronology of a more dominant canon.

\paragraph{Selection Bias}

As the model learns historical time from selected, preserved, and digitized objects that have been made available as data, its predictions must also be interpreted in relation to selection bias.
This is evident in the qualitative results: the most difficult cases tend to have broader prediction intervals, indicating that the model detects a weaker or less stable chronological signal for objects that are visually ambiguous, atypical, or underrepresented (\crefrange{fig:carpaccio-accademia}{fig:de-coxie-anoniem}).
However, wider intervals do not themselves correct the representational asymmetries inherited from the underlying collections or pretrained embedding spaces.
They make uncertainty visible, but do not remove the historical and institutional conditions that produced uneven visibility in the first place.
Therefore, the qualitative analysis supports the paper's broader claim that uncertainty-aware temporal estimation can identify usable chronological structure in pretrained visual representations while also exposing the historically uneven conditions under which such structure becomes learnable.
In this regard, the results corroborate Foka and Griffin's concern that computational systems may exacerbate disparities in the representation and interpretation of cultural artifacts~\cite{foka2024}.

\section{Conclusions and Future Work}
\label{sec:conclusions}

Uncertainty-aware temporal estimation, we have argued, provides a way to examine temporal entanglement at a technical level.
Rather than simply producing more cautious dates, quantile regression and conformal calibration expose the conditions under which historical time becomes learnable from image representations.
However, this process is paradoxical because the method that renders temporality computable can also detach prediction from historical time itself.
Successful predictions identify regions of art-historical visual culture stabilized by dominant genres, media, and institutional afterlives; conversely, failed predictions mark the limits of this stabilization by indicating cases in which style, iconography, and cataloged production dates cease to align.
Therefore, uncertainty-aware temporal estimation should be understood as a method of dating and a critical tool for investigating the historically uneven conditions under which visual temporality becomes computable.

Future work should build upon this diagnostic framework in two areas.
First, the temporal entanglement of more explicitly modeled geographical, institutional, and cultural contexts should be examined. 
Second, image-only estimates should be compared with multimodal approaches that incorporate titles, descriptions, or other metadata.
This would clarify the circumstances in which pretrained representations facilitate historically meaningful temporal inference and those in which they perpetuate the conditions that render cultural artifacts available as data in the first place.


\bibliographystyle{splncs04}
\bibliography{references.bib}

\end{document}